\documentclass{article}

\usepackage[nonatbib,preprint]{nips_2018}

\usepackage[utf8]{inputenc} 
\usepackage[T1]{fontenc}    
\usepackage{hyperref}       
\usepackage{url}            
\usepackage{booktabs}       
\usepackage{amsfonts}       
\usepackage{nicefrac}       
\usepackage{microtype}      
\usepackage{tikz}
\usepackage[english]{babel}

\usepackage[round]{natbib}
\usepackage{verbatim}  
\usepackage{amsmath}
\DeclareMathOperator*{\argmax}{arg\,max}
\DeclareMathOperator*{\argmin}{arg\,min}

\title{Formatting instructions for NIPS 2018}
\title{Artist Style Transfer Via Quadratic Potential}

\author{
  Rahul Bhalley \\
  Department of Information Technology \\
  Guru Nanak Dev Engineering College \\
  Ludhiana, India \\
  \texttt{rahulbhalley@protonmail.com}
  \And
  Jianlin Su \\
  School of Mathematics \\
  Sun Yat-sen University \\
  Guangdong, China \\
  \texttt{bojone@spaces.ac.cn}
}

\begin{document}

\maketitle


\begin{abstract}
In this paper we address the problem of artist style transfer where the painting style of a given artist is applied on a real world photograph. We train our neural networks in adversarial setting via recently introduced quadratic potential divergence for stable learning process. To further improve the quality of generated artist stylized images we also integrate some of the recently introduced deep learning techniques in our method. To our best knowledge this is the first attempt towards artist style transfer via quadratic potential divergence. We provide some stylized image samples in the supplementary material \ref{material}.
The source code for experimentation was written in PyTorch \citep{paszke2017automatic} and is available online in my GitHub repository
\footnote{\url{https://github.com/rahulbhalley/cyclegan-qp}}.
\end{abstract}


\section{Introduction}

In a recent decade deep convolutional neural networks have had a profound effect on the advancement of important computer vision tasks. It has made monumental progress on image classification \citep{krizhevsky2012imagenet}, image generation \citep{goodfellow2014generative}, image style transfer \citep{gatys2015neural, johnson2016perceptual}, music style transfer \citep{huang2018timbretron}, image in-painting \citep{iizuka2017globally, yu2018generative}, image super resolution \citep{chen2017face, wang2018fully}, et cetera. In this paper we are particularly interested in artist style transfer: we reframe our problem as an unsupervised image-to-image translation task where an image is input to the network for transformation into the desired image. We suspect that image-to-image translation problem is \emph{generic} enough to cover some of the previously mentioned tasks. For instance, image in-painting can be framed as an image-to-image translation problem where a small portion of input image is occluded and the task is to recover the desired original image. And image super resolution is a task where input image of low resolution must be filled with the desired information therefore \emph{translating} it into a high resolution image. Similarly, image style transfer \citep{gatys2015neural} may also be reformulated as a translation problem where a real world photograph must be translated into a desired artist style image. 

The vanilla neural image style transfer technique introduced by \citep{gatys2015neural} stylizes a content image based on a style image by iteratively optimizing the noisy image using feature activations of a convolutional network like VGG network \citep{simonyan2014very} pre-trained on ImageNet \citep{russakovsky2015imagenet} challenge dataset to extract the content and style features from content and style images, respectively. Both \citep{johnson2016perceptual, ulyanov2016texture} independently proposed to use convolutional feed forward network for stylizing an image into a desired artist's painting style. Recently \citep{zhu2017unpaired} proposed CycleGAN using a powerful generative modeling framework, generative adversarial network \citep{goodfellow2014generative}, for image translations. It learns a two-way mapping i.e. from domain $\mathbb{A}$ to domain $\mathbb{B}$ and vice-versa using two GANs constrained with a cycle-consistency loss term.

\begin{figure}
  \centering
  \includegraphics[height=2.8cm]{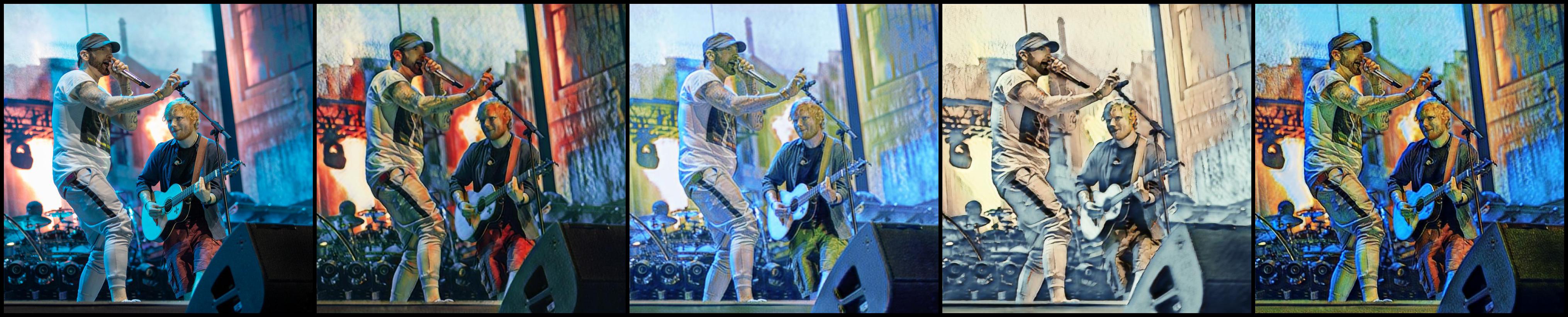}
  \caption{Samples of various artists stylized images (from left to right): a) original image, b) Paul Cézanne, c) Claude Monet, d) Ukiyo-e, and e) Vincent Van Gogh. Please zoom into the images for better visualization.}
  \label{fig:intro}
\end{figure}

The following sections of paper are formatted such that in section \ref{gan_pre} we review state-of-the-art GAN framework from divergence perspective. We present our proposed method in section \ref{cyclegan_plus_plus} following the experiment details in section \ref{expt}. Finally, we present some samples from our network for artist style transfer in supplementary material section \ref{material}.


\section{GAN Preliminaries}
\label{gan_pre}

Generative adversarial network (GAN) is an unsupervised neural network framework for training generative network $\mathcal{G}$ by using the training information from critic network $\mathcal{C}$ by back-propagating the errors from it. $\mathcal{C}$ learns to distinguish between fake samples generated by $\mathcal{G}$ and the real data samples from data distribution $\mathbb{P}_{d}$. And $\mathcal{G}$ learns to generate more realistic samples resembling to those from $\mathbb{P}_{d}$ by using training information from $\mathcal{C}$. Here, both $\mathcal{C}$ and $\mathcal{G}$ are in adversarial situation where $\mathcal{C}$ learns to label the fake and real data correctly while $\mathcal{G}$ tends to learn a way to generate realistic data to convince $\mathcal{C}$ into believing that the generated data is real.Theoretically, at an optimal stage when Nash equilibrium is achieved $\mathbb{P}_g \approx \mathbb{P}_d$ but in practice it is difficult to train a GAN to reach this optimal stage. Many different techniques has been devised \citep{salimans2016improved, arjovsky2017towards} for overcoming the problems of GANs such as mode collapse, gradient vanishing, careful design of architectures for high resolution training, et cetera. But it may be noticed that recently the trend has shifted towards improving the choice of divergence which in turn seems to have been helpful in stabilizing the training of GANs.

In recent years the trend towards shifting from Jensen-Shannen $\mathcal{JS}_{div}$ to Wasserstein $\mathcal{W}_{div}$ divergence for training a GAN framework have considerably improved the training process. But for constraining $\mathcal{C}$ to 1-Lipschitz set of functions, for creating Wasserstein GANs, has led to increasing complexity in loss function formulation of $\mathcal{C}$. Recently to suppress this complexity \citep{su2018gan} proposed quadratic potential divergence $\mathcal{Q}_{div}$ and showed that their formulation of min-max game have better properties than those of $\mathcal{JS}_{div}$ or $\mathcal{W}_{div}$ divergences based GANs \citep{goodfellow2014generative, arjovsky2017wasserstein, gulrajani2017improved, salimans2018improving, wei2018improving, wu2017wasserstein}. Also note that although GANs introduced by \citep{brock2018large, karras2017progressive, karras2018style, salimans2018improving} improves the quality of image generation considerably but at a cost of very high computation resources which are usually infeasible to acquire.

Notably $\mathcal{Q}_{div}$ have two desired benefits: a) it retains the {\em no constraint on $\mathcal{C}$} characteristic of standard GAN (Goodfellow et al., 2014) and b) it {\em eliminates the gradient vanishing problem} without till yet increasing complexity of empirical tricks for constraining $\mathcal{C}$ to lie in 1-Lipschitz set of functions for constructing Wasserstein GAN \citep{arjovsky2017wasserstein}. Moreover \citep{su2018gan} also showed that $\mathcal{Q}_{div}$ behaves as if it already has adaptive Lipschitz constraint property present in it (for details refer section B.2 of \citep{su2018gan}). The formulation of quadratic potential divergence is as follows:
\begin{equation}
\label{eq:q_div_eq}
\mathcal{Q}_{div}(x_{r}, x_{f}) = a - \frac{a^{2}}{2\lambda d(x_{r}, x_{f})}
\end{equation}

where $a = \mathcal{C}(x_{r}, x_{f}) - \mathcal{C}(x_{f}, x_{r})$, $\lambda > 0$ is a hyper-parameter, $x_{r} \sim \mathbb{P}_{d}$ is a real data sample, $x_{f} \sim \mathbb{P}_{g}$ is a fake data sample generated by $\mathcal{G}$ with prior on latent vector $z \sim \mathbb{P}_{z}$ Gaussian distribution. Note that $\mathcal{C} : x \rightarrow [0, 1]$ is a differentiable critic function where $x \in \{x_{r}, x_{f}\}$. Here, $d(x_{r}, x_{f})$ is the distance metric between samples $x_{r}$ and $x_{f}$ which we choose in our experiments to be the $L_{1}$ distance metric:
\begin{equation}
d(x_{r}, x_{f}) = \|x_{r} - x_{f}\|_{1}
\end{equation}

Note that (\ref{eq:q_div_eq}) may also be viewed as a \emph{quadratic function} which has no constraint on $\mathcal{C}$ like standard GAN and exhibits the properties similar to Wasserstein GAN. To train a GAN with $\mathcal{Q}_{div}$ the following adversarial loss functions arise:
\begin{equation}
\mathcal{C} = \argmax_{\mathcal{C}} \mathbb{E}_{(x_{r}, x_{f}) \sim \mathbb{P}_{r}\mathbb{P}_{f}} [\mathcal{Q}_{div}(x_{r}, x_{f})]
\end{equation}
\begin{equation}
\mathcal{G} = \argmin_{\mathcal{G}} \mathbb{E}_{(x_{r}, x_{f}) \sim \mathbb{P}_{r}\mathbb{P}_{f}} [\mathcal{C}(x_{r}, x_{f}) - \mathcal{C}(x_{f}, x_{r})]
\end{equation}

Due to the simplicity and faster computation of $\mathcal{Q}_{div}$ than the time consuming approximation of $\mathcal{W}_{div}$ we choose to use $\mathcal{Q}_{div}$ formulation to play min-max game between our adversarial networks in our work for training $\mathcal{G}$ to stylize image data. Note that \citep{zhu2017unpaired} used least squares GAN \citep{mao2017least} in their work. 


\section{CycleGAN-QP}
\label{cyclegan_plus_plus}

In unsupervised image-to-image translation problem the task is to learn a generator/translation function $\mathcal{G}^{rs} : \mathbb{D}^{r} \rightarrow \mathbb{D}^{s}$ mapping \emph{image} samples from domain $\mathbb{D}^{r}$ to $\mathbb{D}^{s}$. We assume datasets of $\mathbb{D}^{r}$ and $\mathbb{D}^{s}$ domains to be unpaired in terms of similarity of images. For the mapping between both domains to be meaningful the bijection property of functions is imposed by learning an inverse differentiable function $\mathcal{G}^{sr} : \mathbb{D}^{s} \rightarrow \mathbb{D}^{r}$ and cycle-consistency functions: a) $\mathcal{G}^{rs}(\mathcal{G}^{sr}(x^{s})) \approx x^{s}$, and b) $\mathcal{G}^{sr}(\mathcal{G}^{rs}(x^{r})) \approx x^{r}$ where $x^{r} \sim \mathbb{D}^{r}$, $x^{s} \sim \mathbb{D}^{s}$.

We made some modifications to the previous of CycleGAN. First, we use the recently introduced \emph{quadratic potential} divergence $\mathcal{Q}_{div}$ \citep{su2018gan} for adversarial training of our proposed \emph{CycleGAN-QP}. The empirical explanation for using this divergence over $\mathcal{JS}_{div}$ and $\mathcal{W}_{div}$ is twofold: a) it retains the simplicity of standard GAN \citep{goodfellow2014generative} and b) it has properties similar to or even better than Wasserstein GAN \citep{arjovsky2017wasserstein} such as adaptive Lipschitz constraint which helps in utilizing the representation capacity of $\mathcal{C}$ more effectively rather than under-utilizing it by constraining it to a subspace of 1-Lipschitz set of functions like in case of Wasserstein GAN. 

Inspired by the work of \citep{shrivastava2017learning} we also use the identity loss term in addition to cycle-consistency loss term in generators to enforce the generated image to be similar to the original image in terms of overall structure of content. Note that identity loss term was also used in original formulation \citep{zhu2017unpaired}. Both the generators requires three losses in our training formulation of method: a) adversarial quadratic potential $\mathcal{Q}_{div}$ loss term $\mathcal{G}_{\mathcal{Q}}$, b) cyclic-consistency loss term $\mathcal{G}_{cyc}$, c) and identity loss term $\mathcal{G}_{id}$ written as follows for each domain translational $\mathcal{G}$:

\begin{equation}
\label{eq:adv_loss_1}
\mathcal{G}_{\mathcal{Q}}^{rs} = \mathbb{E}_{x^{s} = \mathcal{G}^{rs}(x^{r}), x^{r} \sim \mathbb{D}^{r}} [\mathcal{C}(x^{r}, x^{s}) - \mathcal{C}(x^{s}, x^{r})]
\end{equation}
\begin{equation}
\label{eq:adv_loss_2}
\mathcal{G}_{\mathcal{Q}}^{sr} = \mathbb{E}_{x^{r} = \mathcal{G}^{sr}(x^{s}), x^{s} \sim \mathbb{D}^{s}} [\mathcal{C}(x^{s}, x^{r}) - \mathcal{C}(x^{r}, x^{s})]
\end{equation}
\begin{equation}
\label{eq:cyc_loss_1}
\mathcal{G}_{cyc}^{r} = \mathbb{E}_{x^{r} \sim \mathbb{D}^{r}}\|\mathcal{G}^{sr}(\mathcal{G}^{rs}(x^{r})) - x^{r}\|_{1}
\end{equation}
\begin{equation}
\label{eq:cyc_loss_2}
\mathcal{G}_{cyc}^{s} = \mathbb{E}_{x^{s} \sim \mathbb{D}^{s}}\|\mathcal{G}^{rs}(\mathcal{G}^{sr}(x^{s})) - x^{s}\|_{1}
\end{equation}
\begin{equation}
\label{eq:id_loss_1}
\mathcal{G}_{id}^{r} = \mathbb{E}_{(x^{r}, x^{s}) \sim \mathbb{D}^{r}\mathbb{D}^{s}}\|\mathcal{G}^{sr}(x^{s}) - x^{r}\|_{1}
\end{equation}
\begin{equation}
\label{eq:id_loss_2}
\mathcal{G}_{id}^{s} = \mathbb{E}_{(x^{r}, x^{s}) \sim \mathbb{D}^{r}\mathbb{D}^{s}}\|\mathcal{G}^{rs}(x^{r}) - x^{s}\|_{1}
\end{equation}

The adversarial $\mathcal{Q}_{div}$ losses are represented by equations (\ref{eq:adv_loss_1}), (\ref{eq:adv_loss_2}) above and (\ref{eq:cyc_loss_1}), (\ref{eq:cyc_loss_2}) are cyclic-consistency loss terms for $\mathbb{D}^{r}$ and $\mathbb{D}^{s}$ respectively. Similarly, equations (\ref{eq:id_loss_1}) and (\ref{eq:id_loss_2}) represents losses for preserving the identity of samples from respective $\mathbb{D}^{r}$ and $\mathbb{D}^{s}$. Therefore, the overall equation for generators to optimize distills down to the following equation:

\begin{equation}
\begin{split}
\label{eq:g_min}
\mathcal{G}
= \argmin_{\mathcal{G}}\Big[
\mathbb{E}_{(x^{r}, x^{s}) \sim \mathbb{D}^{r}\mathbb{D}^{s}}\big[\mathcal{G}_{\mathcal{Q}}^{rs}(x^{r}, x^{s}) + \mathcal{G}_{\mathcal{Q}}^{sr}(x^{r}, x^{s})\big] + \\
\alpha \big[\mathbb{E}_{x^{r} \sim \mathbb{D}^{r}}[\mathcal{G}_{cyc}^{r}(x^{r})] + \mathbb{E}_{x^{s} \sim \mathbb{D}^{s}}[\mathcal{G}_{cyc}^{s}(x^{s})]\big] + \\
\beta \big[\mathbb{E}_{(x^{r}, x^{s}) \sim \mathbb{D}^{r}\mathbb{D}^{s}}[\mathcal{G}_{id}^{r}(x^{r}, x^{s}) + \mathcal{G}_{id}^{s}(x^{r}, x^{s})\big]
\Big]
\end{split}
\end{equation}

where, $\alpha$ and $\beta$ are hyper-parameters to weight the importance of cycle-consistency and identity loss terms. 

Next, the complete adversarial equation for critic networks to optimize is as follows:

\begin{equation}
\begin{split}
\mathcal{C}
= \argmax_{\mathcal{C}}
\Bigg[
\mathbb{E}_{(x^{r}, x^{s}) \sim \mathbb{D}^{r}\mathbb{D}^{s}}\bigg(\mathcal{C}^{rs}(x^{r}, x^{s}) - \mathcal{C}^{rs}(x^{s}, x^{r}) - \frac{(\mathcal{C}^{rs}(x^{r}, x^{s}) - \mathcal{C}^{rs}(x^{s}, x^{r}))^{2}}{2 \lambda \|x^{r} - x^{s}\|_{1}}\bigg) + 
\\
\mathbb{E}_{(x^{r}, x^{s}) \sim \mathbb{D}^{r}\mathbb{D}^{s}}\bigg(\mathcal{C}^{sr}(x^{r}, x^{s}) - \mathcal{C}^{sr}(x^{s}, x^{r}) - \frac{(\mathcal{C}^{sr}(x^{r}, x^{s}) - \mathcal{C}^{sr}(x^{s}, x^{r}))^{2}}{2 \lambda \|x^{r} - x^{s}\|_{1}}\bigg)
\Bigg]
\end{split}
\end{equation}

Finally, a problem where the images generated by convolutional neural networks has some checkerboard artifacts was recently put under light by \citep{odena2016deconvolution}. It was shown that when training transpose convolution layers especially for two dimensional data these artifacts are intensified which the transpose convolution itself is unable to remove automatically by learning its filters. They proposed to substitute transpose convolution with the interpolation of input with nearest neighbor following convolution operation to eliminate the checkerboard artifacts. More recently \citep{donahue2018synthesizing} successfully used it to eliminate noise from raw audio generated by GANs. Similarly, we also adopt this technique to improve the resulting stylized images and we find no checkerboard artifacts in our generated art pieces and reconstructed original images.


\section{Experiments}
\label{expt}

In this section we present the experimental details. 

\paragraph{Dataset}
We used the same painting datasets collected by \citep{zhu2017unpaired} in our experiments. Various artists' paintings were trained for style transfer namely, a) Paul Cézanne, b) Claude Monet, c) Ukiyo-e, and d) Vincent Van Gogh. The stylized images in Figures \ref{fig:sty} and \ref{fig:rec} were scrapped manually from Pexels website\footnote{\url{https://www.pexels.com}} while the image of Eminem performing River song with Ed Sheeran in Figure \ref{fig:intro} was scrapped from Southpawer website\footnote{\url{https://www.southpawer.com}}.

\paragraph{Data Preprocessing}
For training we first stochastically flipped the original images across the vertical axis i.e. horizontally following the center-cropping of these input images to $256^{2}$ resolution. Finally the images were normalized with a mean and standard deviation of $0.5$ each for all the channels. During training the images were sampled randomly for stochastically training our CycleGAN-QP. 

\paragraph{Neural Network}
We designed our network similar to \citep{zhu2017unpaired} but it varies by some minor modifications. Our generator network $\mathcal{G}$ is composed of three subnetworks namely, a) encoder, b) transformer, and c) decoder where transformer has residual connections \citep{he2016deep}. And critic $\mathcal{C}$ is a fully convolutional network. The parameters of all layers were initialized with Glorot technique \citep{glorot2010understanding} using uniform distribution. More details can be found in our implementation. 

\paragraph{Training}
In all the experiments a batch size of $4$ was used for optimizing the networks with stochastic gradient descent. Specifically, the networks were trained with Adam \citep{kingma2014adam} optimizer with a constant learning rate of $0.0002$ throughout training. Adam's hyper-parameter configurations $\beta_{1}$ and $\beta_{2}$ were tuned to $0.5$ and $0.999$ values respectively. Note that in equation (\ref{eq:q_div_eq}) we set $\lambda$ to be $10$ and the values of $\alpha$ and $\beta$ in equation (\ref{eq:g_min}) were set to be $10$ and $0.5$ respectively. We trained the network for $15000$ iterations (except the networks for Ukiyo-e dataset were trained for up to $12000$ iterations) and the average time taken is nearly $1.25$ days on a single NVIDIA K80 GPU machine. The reason behind setting a batch size of $4$ is our limited access to GPU resources. 


\section{Conclusion}

In this paper we introduced our method CycleGAN-QP built upon the two major improvements for unsupervised artist style transfer and reconstruction between unpaired datasets: a) quadratic potential divergence for simplified and more stable state-of-the-art generative adversarial training than the standard and Wasserstein GANs, and b) removal of checkerboard artifacts. We conclude that our method performs reasonably good quality artist style transfer (see Figure \ref{fig:sty}) and its reconstruction back to original image (see Figure \ref{fig:rec}).


\subsection*{Acknowledgements}

We would like to thank the PyTorch developer team for creating an amazing neural network library. 


\bibliographystyle{plainnat}
\bibliography{nips_2018}


\newpage

\section{Supplementary Material}
\label{material}

In following Figures \ref{fig:sty} and \ref{fig:rec}, the samples of various artists stylized and recovered images (from left to right) are shown: a) original image, b) Paul Cézanne, c) Claude Monet, d) Ukiyo-e, and e) Vincent Van Gogh. All of these images were sampled at $1024^{2}$ resolution. 

\begin{figure}[h]
  \centering
  \includegraphics[height=12cm]{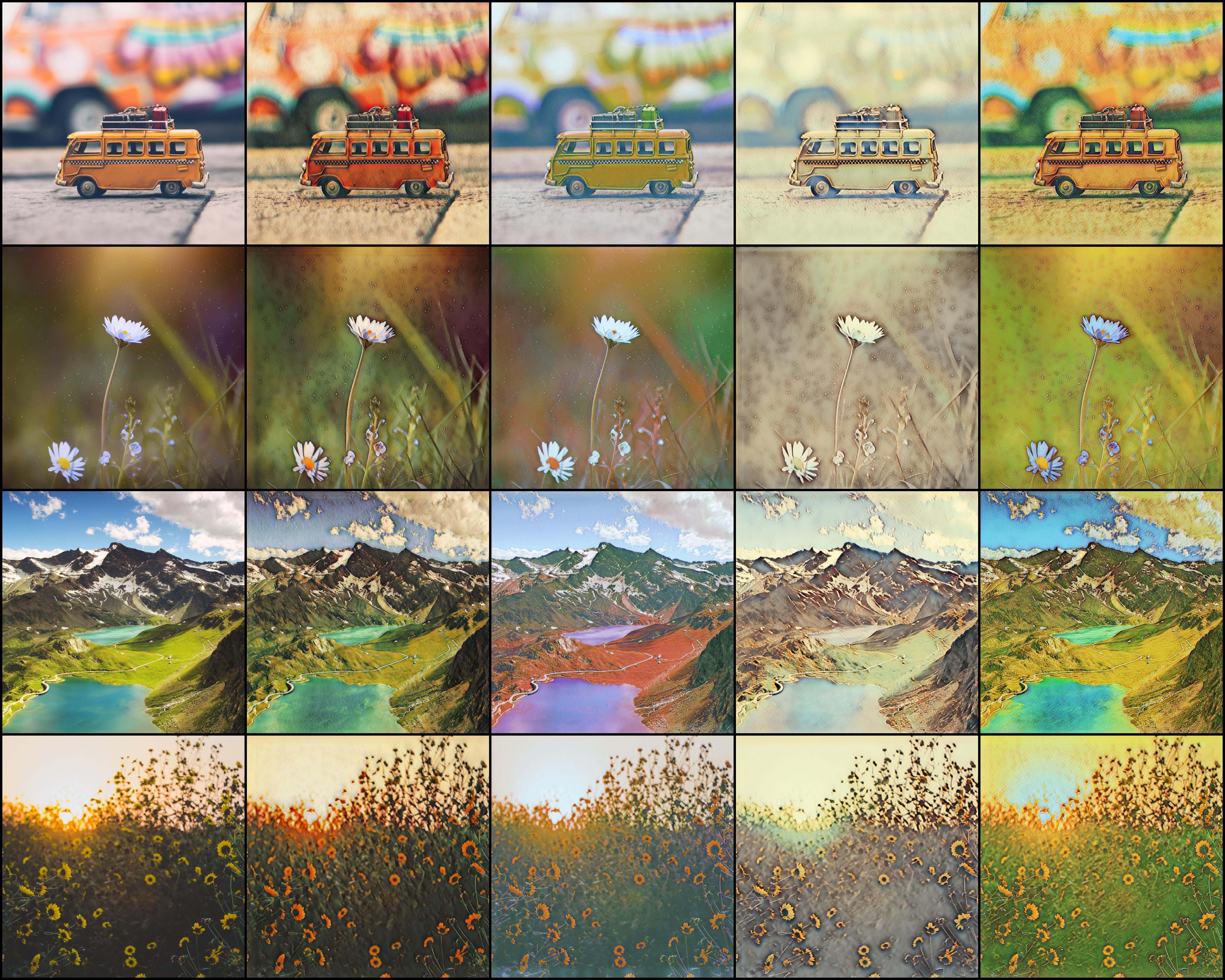}
  \caption{Artists stylized images.}
  \label{fig:sty}
\end{figure}

\begin{figure}[h]
  \centering
  \includegraphics[height=12cm]{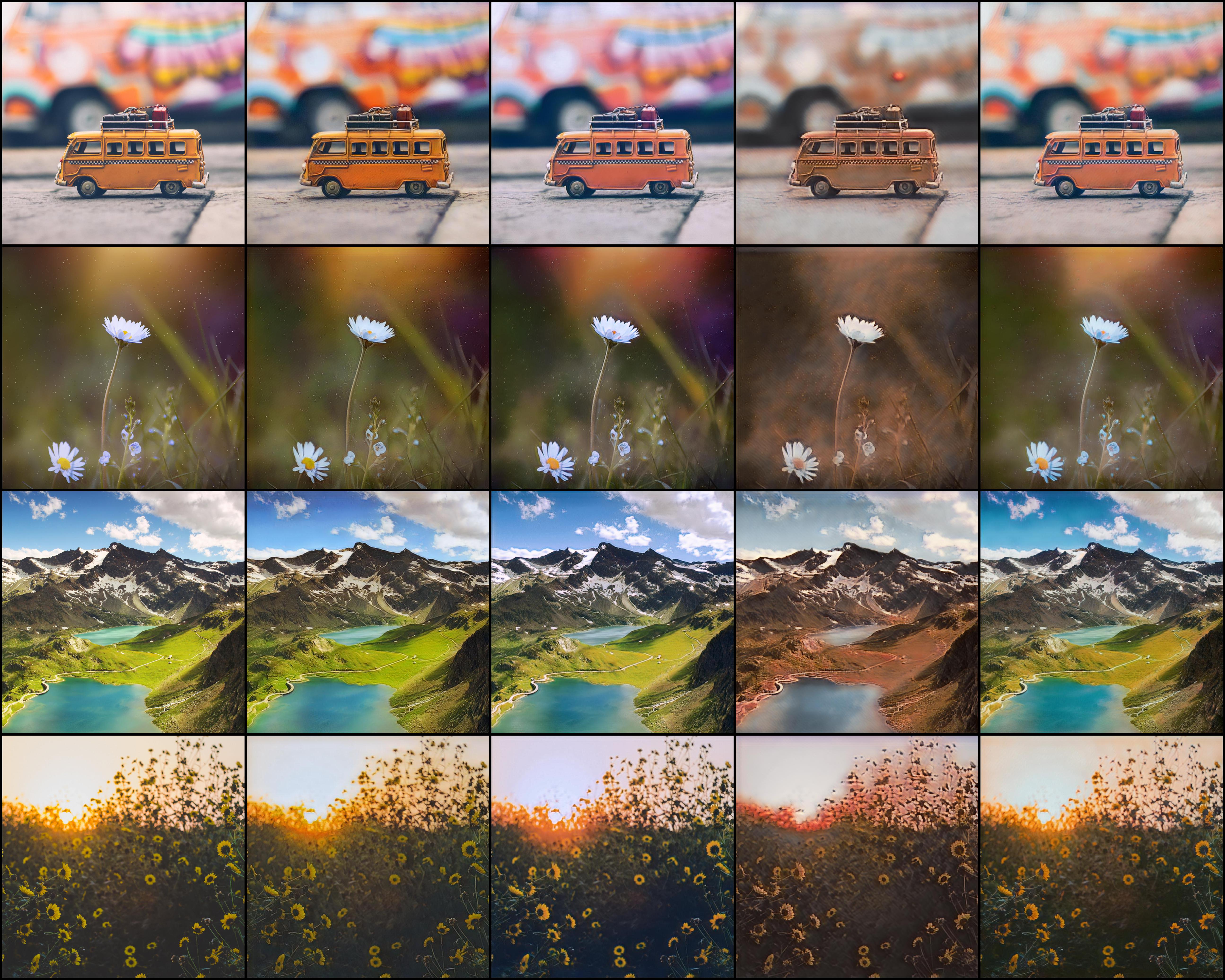}
  \caption{Recovered images from artist stylized real images.}
  \label{fig:rec}
\end{figure}

\end{document}